\documentclass[runningheads,a4paper]{llncs}

\usepackage{amsmath}
\usepackage{amssymb}
\usepackage{graphicx}
\usepackage{caption}
\usepackage{subcaption}
\usepackage{multirow}

\usepackage{xcolor}
\usepackage{hyperref}

\newcommand{\species}[1]{\textit{#1}}

\usepackage{url}
\urldef{\mailsa}\path|cicek@cs.uni-freiburg.de|
\urldef{\mailsb}\path|olafr@google.com|
\urldef{\mailsc}\path|soeren.lienkamp@uniklinik-freiburg.de|  
\newcommand{\keywords}[1]{\par\addvspace\baselineskip
\noindent\keywordname\enspace\ignorespaces#1}

\begin{document}

\mainmatter

\title{3D U-Net: Learning Dense Volumetric Segmentation from Sparse Annotation}

\titlerunning{Volumetric Segmentation with the 3D U-Net}

\author{\"{O}zg\"{u}n \c{C}i\c{c}ek \inst{1,2} 
\and    Ahmed Abdulkadir \inst{1,4} 
\and    Soeren S. Lienkamp \inst{2,3} 
\and    Thomas Brox \inst{1,2} 
\and    Olaf Ronneberger \inst{1,2,5}} 
\authorrunning{Volumetric Segmentation with the 3D U-Net}

\institute{Computer Science Department, University of Freiburg, Germany
\and BIOSS Centre for Biological Signalling Studies, Freiburg, Germany
\and University Hospital Freiburg, Renal Division, Faculty of Medicine, University of Freiburg, Germany
\and Department of Psychiatry and Psychotherapy, University Medical Center Freiburg, Germany
\and Google DeepMind, London, UK\\
\mailsa\\}

\toctitle{Lecture Notes in Computer Science}
\tocauthor{Authors' Instructions}
\maketitle

\setcounter{footnote}{0}

\begin{abstract}
This paper introduces a network for volumetric segmentation that learns from sparsely annotated volumetric images. We outline two attractive use cases of this method: (1) In a semi-automated setup, the user annotates some slices in the volume to be segmented. The network learns from these sparse annotations and provides a dense 3D segmentation. (2) In a fully-automated setup, we assume that a representative, sparsely annotated training set exists. Trained on this data set, the network densely segments new volumetric images. The proposed network extends the previous u-net architecture from Ronneberger et al. by replacing all 2D operations with their 3D counterparts. The implementation performs on-the-fly elastic deformations for efficient data augmentation during training. It is trained end-to-end from scratch, i.e., no pre-trained network is required. We test the performance of the proposed method on a complex, highly variable 3D structure, the \species{Xenopus} kidney, and achieve good results for both use cases.

\end{abstract}

\section{Introduction}

\begin{figure}

\includegraphics[width=\textwidth]{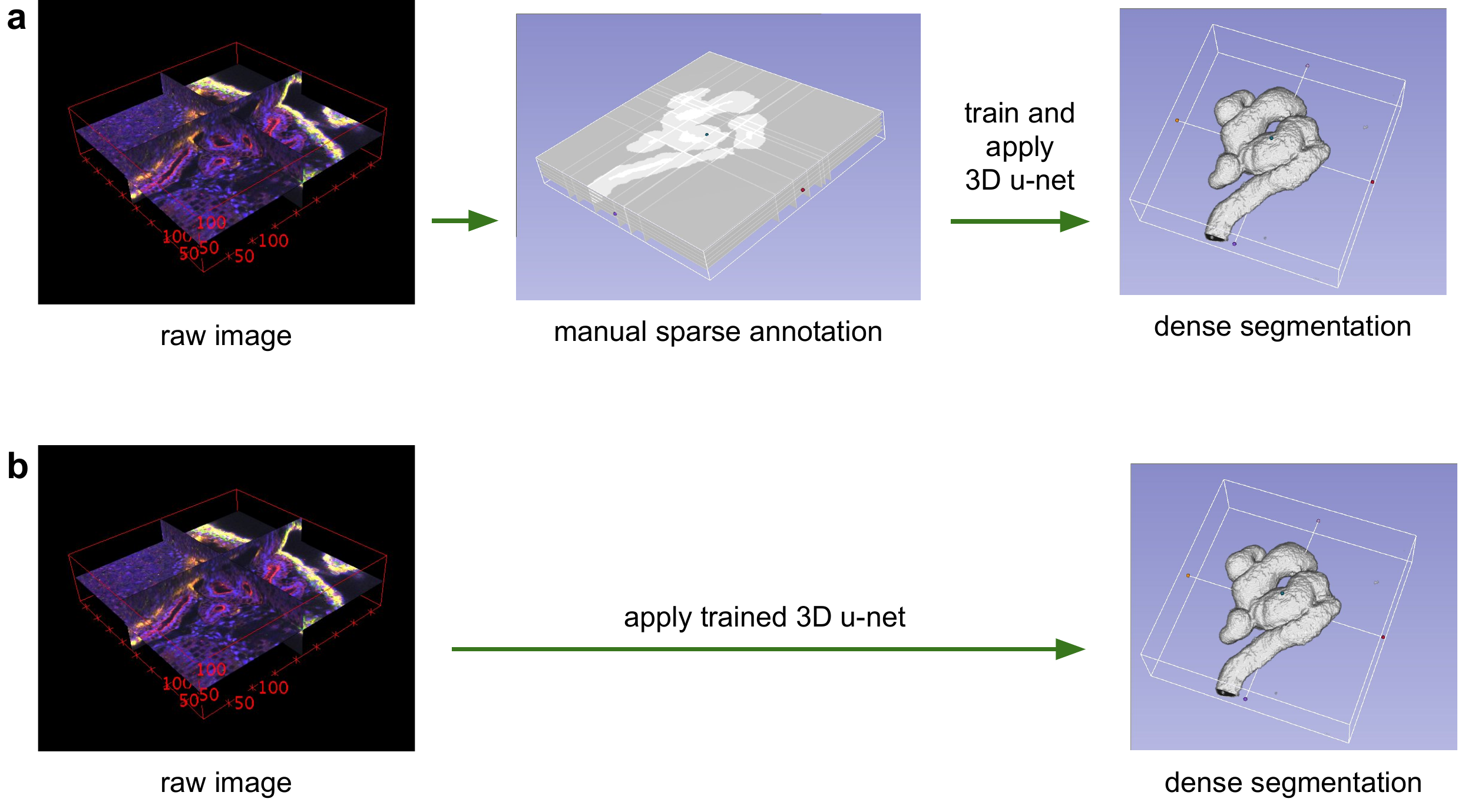}
\caption{Application scenarios for volumetric segmentation with the 3D u-net. (a) Semi-automated segmentation: the user annotates some slices of each volume to be segmented. The network predicts the dense segmentation. (b) Fully-automated segmentation: the network is trained with annotated slices from a representative training set and can be run on non-annotated volumes.}
\label{fig:scenario}
\end{figure}

Volumetric data is abundant in biomedical data analysis. Annotation of such data with segmentation labels causes difficulties, since only 2D slices can be shown on a computer screen. Thus, annotation of large volumes in a slice-by-slice manner is very tedious. It is inefficient, too, since neighboring slices show almost the same information. Especially for learning based approaches that require a significant amount of annotated data, full annotation of 3D volumes is not an effective way to create large and rich training data sets that would generalize well. 

In this paper, we suggest a deep network that learns to generate dense volumetric segmentations, but only requires some annotated 2D slices for training. This network can be used in two different ways as depicted in Fig.~\ref{fig:scenario}: the first application case just aims on densification of a sparsely annotated data set; the second learns from multiple sparsely annotated data sets to generalize to new data. Both cases are highly relevant. 

The network is based on the previous u-net architecture, which consists of a contracting encoder part to analyze the whole image and a successive expanding decoder part to produce a full-resolution segmentation~\cite{unet}. While the u-net is an entirely 2D architecture, the network proposed in this paper takes 3D volumes as input and processes them with corresponding 3D operations, in particular, 3D convolutions, 3D max pooling, and 3D up-convolutional layers. 
Moreover, we avoid bottlenecks in the network architecture \cite{google} and use batch normalization \cite{bn} for faster convergence.

In many biomedical applications, only very few images are required to train a network that generalizes reasonably well. This is because each image already comprises repetitive structures with corresponding variation. In volumetric images, this effect is further pronounced, such that we can train a network on just two volumetric images in order to generalize to a third one. A weighted loss function and special data augmentation enable us to train the network with only few manually annotated slices, i.e., from sparsely annotated training data.  

We show the successful application of the proposed method on difficult confocal microscopic data set of the \species{Xenopus} kidney. During its development, the \species{Xenopus} kidney forms a complex structure \cite{soeren} which limits the applicability of pre-defined parametric models.
First we provide qualitative results to demonstrate the quality of the densification from few annotated slices. These results are supported by quantitative evaluations. We also provide experiments which shows the effect of the number of annotated slices on the performance of our network. The Caffe\cite{caffe} based network implementation is provided as OpenSource\footnote{\url{http://lmb.informatik.uni-freiburg.de/resources/opensource/unet.en.html}}. 

\subsection{Related Work}

Challenging biomedical 2D images can be segmented with an accuracy close to human performance by CNNs today \cite{unet,2dnet1,2dnet2}. Due to this success, several attempts have been made to apply 3D CNNs on biomedical volumetric data. Milletari et al. \cite{hough-cnn} present a CNN combined with a Hough voting approach for 3D segmentation. However, their method is not end-to-end and only works for compact blob-like structures. The approach of Kleesiek et al. \cite{deepmri} is one of few end-to-end 3D CNN approaches for 3D segmentation. However, their network is not deep and has only one max-pooling after the first convolutions; therefore, it is unable to analyze structures at multiple scales. 
Our work is based on the 2D u-net \cite{unet} which won several international segmentation and tracking competitions in 2015. The architecture and the data augmentation of the u-net allows learning models with very good generalization performance from only few annotated samples. It exploits the fact that properly applied rigid transformations and slight elastic deformations still yield biologically plausible images. 
Up-convolutional architectures like the fully convolutional networks for semantic segmentation \cite{fcn} and the u-net are still not wide spread and we know of only one attempt to generalize such an architecture to 3D \cite{tran}. In this work by Tran et al., the architecture is applied to videos and full annotation is available for training. The highlight of the present paper is that it can be trained from scratch on sparsely annotated volumes and can work on arbitrarily large volumes due to its seamless tiling strategy.

\section{Network Architecture}

\begin{figure}
\centering
\includegraphics[height=5.8cm]{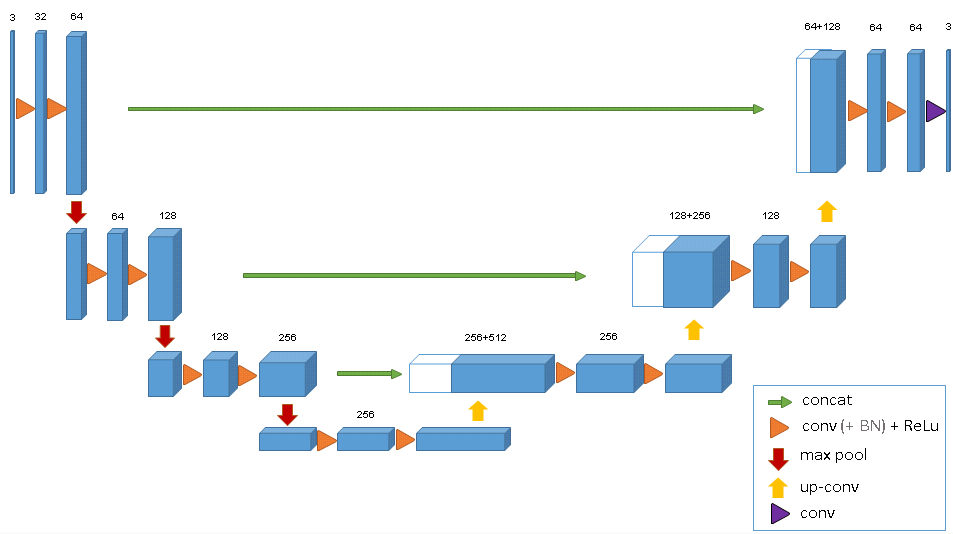}
\caption{The 3D u-net architecture. Blue boxes represent feature maps. The number of channels is denoted above each feature map.}
\label{fig:unet3d}
\end{figure}

Figure~\ref{fig:unet3d} illustrates the network architecture. Like the standard u-net, it has an analysis and a synthesis path each with four resolution steps. In the analysis path, each layer contains two $3 \times 3 \times 3$ convolutions each followed by a rectified linear unit (ReLu), and then a $2 \times 2 \times 2$ max pooling with strides of two in each dimension. In the synthesis path, each layer consists of an upconvolution of $2 \times 2 \times 2$ by strides of two in each dimension, followed by two $3 \times 3 \times 3$ convolutions each followed by a ReLu. Shortcut connections from layers of equal resolution in the analysis path provide the essential high-resolution features to the synthesis path. 
In the last layer a $1 \times 1 \times 1$ convolution reduces the number of output channels to the number of labels which is 3 in our case. The architecture has 19069955 parameters in total.
Like suggested in \cite{google} we avoid bottlenecks by doubling the number of channels already before max pooling. We also adopt this scheme in the synthesis path. 

The input to the network is a $132 \times 132 \times 116$ voxel tile of the image with 3 channels. Our output in the final layer is $44 \times 44 \times 28$ voxels in x, y, and z directions respectively. With a voxel size of $1.76 \times 1.76 \times 2.04 \mu \text{m}^3$, the approximate receptive field becomes $155 \times 155 \times 180 \mu\text{m}^3$ for each voxel in the predicted segmentation. Thus, each output voxel has access to enough context to learn efficiently. 

We also introduce batch normalization (``BN'') before each ReLU. In \cite{bn}, each batch is normalized during training with its mean and standard deviation and global statistics are updated using these values. This is followed by a layer to learn scale and bias explicitly. At test time, normalization is done via these computed global statistics and the learned scale and bias. However, we have a batch size of one and few samples. In such applications, using the current statistics also at test time works the best. 

The important part of the architecture, which allows us to train on sparse annotations, is the weighted softmax loss function. Setting the weights of unlabeled pixels to zero makes it possible to learn from only the labelled ones and, hence, to generalize to the whole volume. 

\section{Implementation Details}

\subsection{Data}

We have three samples of \species{Xenopus} kidney embryos at Nieuwkoop-Faber stage 36-37 \cite{stage}. One of them is shown in Fig.~\ref{fig:scenario} (left). 3D Data have been recorded in four tiles with three channels at a voxel size of $0.88 \times 0.88 \times 1.02 \mu\text{m}^3$ using a Zeiss LSM 510 DUO inverted confocal microscope equipped with a Plan-Apochromat 40x/1.3 oil immersion objective lens. We stitched the tiles to large volumes using XuvTools \cite{xuv}. The first channel shows Tomato-Lectin coupled to Fluorescein at 488nm excitation wavelength. The second channel shows DAPI stained cell nuclei at 405 nm excitation. The third channel shows Beta-Catenin using a secondary antibody labelled with Cy3 at 564nm excitation marking the cell membranes. We manually annotated some orthogonal xy, xz, and yz slices in each volume using Slicer3D\footnote{\url{https://www.slicer.org}} \cite{slicer3d}. The annotation positions were selected according to good data representation i.e. annotation slices were sampled as uniformly as possible in all 3 dimensions. Different structures were given the labels 0: ``inside the tubule''; 1: ``tubule''; 2: ``background'', and 3: ``unlabeled''. All voxels in the unlabelled slices also get the label 3 (``unlabeled''). We ran all our experiments on down-sampled versions of the original resolution by factor of two in each dimension. Therefore, the data sizes used in the experiments are $248 \times 244 \times 64$, $245 \times 244 \times 56$ and $246 \times 244 \times 59$ in $x \times y \times z$ dimensions for our sample 1, 2, and 3, respectively. The number of manually annotated slices in orthogonal (yz, xz, xy) slices are (7, 5, 21), (6, 7, 12), and (4, 5, 10) for sample 1, 2, and 3, respectively. 

\subsection{Training}

Besides rotation, scaling and gray value augmentation, we apply a smooth dense deformation field on both data and ground truth labels. For this, we sample random vectors from a normal distribution with standard deviation of 4 in a grid with a spacing of $32$ voxels in each direction and then apply a B-spline interpolation. The network output and the ground truth labels are compared using softmax with weighted cross-entropy loss, where we reduce weights for the frequently seen background and increase weights for the inner tubule to reach a balanced influence of tubule and background voxels on the loss. Voxels with label 3 (``unlabled'') do not contribute to the loss computation, i.e. have a weight of 0. We use the stochastic gradient descent solver of the Caffe \cite{caffe} framework for network training. To enable training of big 3D networks we used the memory efficient cuDNN\footnote{\url{https://developer.nvidia.com/cudnn}} convolution layer implementation. Data augmentation is done on-the-fly, which results in as many different images as training iterations. We ran 70000 training iterations on an NVIDIA TitanX GPU, which took approximately 3 days.

\section{Experiments}

\subsection{Semi-Automated Segmentation}\label{subsec:general}

For semi-automated segmentation, we assume that the user needs a full segmentation of a small number of volumetric images, and does not have prior segmentations. The proposed network allows the user to annotate a few slices from each volume and let the network create the dense volumetric segmentation. 

\begin{figure}
\centering
\begin{subfigure}{.5\textwidth}
  \centering
  \includegraphics[height=3.7cm]{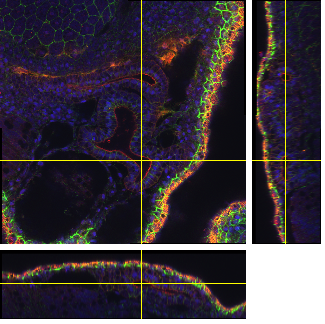}
  \caption{}
  \label{fig:gen1}
\end{subfigure}%
\begin{subfigure}{.5\textwidth}
  \centering
  \includegraphics[height=3.7cm]{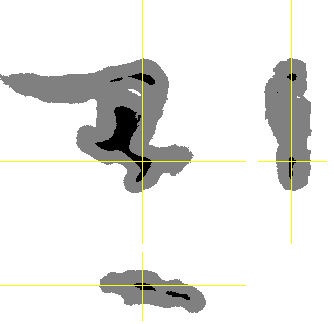}
  \caption{}
  \label{fig:gen2}
\end{subfigure}
\caption{\textbf{(a)} The confocal recording of our 3rd \species{Xenopus} kidney. \textbf{(b)} Resulting dense segmentation from the proposed 3D u-net with batch normalization.}
\label{fig:gen}
\end{figure}

For a \textit{qualitative} assessment, we trained the network on all three sparsely annotated samples. Figure~\ref{fig:gen} shows the segmentation results for our 3rd sample. The network can find the whole 3D volume segmentation from a few annotated slices and saves experts from full volume annotation.

To assess the \textit{quantitative} performance in the semi-automated setup, we uniformly partitioned the set of all 77 manually annotated slices from all 3 samples into three subsets and did a 3-fold cross validation both with and without batch normalization. To this end, we removed the test slices and let them stay unlabeled. This simulates an application where the user provides an even sparser annotation. To measure the gain from using the full 3D context, we compare the results to a pure 2D implementation, that treats all labelled slices as independent images. We present the results of our experiment in Table~\ref{tab:cv_results}. Intersection over Union (IoU) is used as accuracy measure to compare dropped out ground truth slices to the predicted 3D volume. The IoU is defined as \textit{true positives/(true positives + false negatives + false positives)}. The results show that our approach is able to already generalize from only very few annotated slices leading to a very accurate 3D segmentation with little annotation effort.

We also analyzed the effect of the number of annotated slices on the network performance. To this end, we simulated a one sample semi-automated segmentation. We started using 1 annotated slice in each orthogonal direction and increased the number of annotated slices gradually. We report the high performance gain of our network for each sample (S1, S2, and S3) with every few additional ground truth (``GT'') slices in Table~\ref{tab:numofslices}. The results are taken from networks which were trained for 10 hours with batch normalization. For testing we used the slices not used in any setup of this experiment.

\begin{table}
\begin{minipage}{0.45\textwidth}
\centering
\caption{Cross validation results for semi-automated segmentation (IoU)}

\begin{tabular}{c@{~~}c@{~~}c@{~~}c} 
\hline
test & 3D & 3D  & 2D  \\
slices & w/o BN & with BN & with BN \\
\hline
subset 1 & 0.822 & 0.855 &  0.785  \\
subset 2 & 0.857 & 0.871 &  0.820  \\ 
subset 3 & 0.846 & 0.863 &  0.782  \\ \hline
average & 0.842 & 0.863 &  0.796  \\ 
\hline
\end{tabular}
\label{tab:cv_results}
\end{minipage}
\hfill
\begin{minipage}{0.45\textwidth}
\centering
\caption{Effect of \# of slices for semi-automated segmentation (IoU)}

\label{tab:numofslices}
\begin{tabular}{c@{~~}c@{~~}c@{~~}c@{~~}c} 
\hline
  GT    &   GT   & IoU & IoU & IoU \\
slices  & voxels & S1  & S2  & S3  \\
\hline
1,1,1 & 2.5\% & 0.331 & 0.483 & 0.475\\
2,2,1 & 3.3\% & 0.676 & 0.579 & 0.738\\ 
3,3,2 & 5.7\% & 0.761 & 0.808 & 0.835\\
5,5,3 & 8.9\% & 0.856 & 0.849 & 0.872\\
\hline
\end{tabular}
\end{minipage}
\end{table}
\begin{table}
\centering
\caption{Cross validation results for fully-automated segmentation (IoU)}

\label{tab:cv_results_naive}
\begin{tabular}{c@{~~~}c@{~~~}c@{~~~}c} 
\hline
test & 3D& 3D  & 2D \\
volume &  w/o BN   & with BN  & with BN  \\
\hline
1 & 0.655 & 0.761 & 0.619\\
2 & 0.734 & 0.798 & 0.698\\ 
3 & 0.779 & 0.554 & 0.325\\ \hline
average & 0.723 & 0.704 & 0.547\\
\hline
\end{tabular}
\end{table}

\subsection{Fully-automated Segmentation}

The fully-automated segmentation setup assumes that the user wants to segment a large number of images recorded in a comparable setting. We further assume that a representative training data set can be assembled. 

To estimate the performance in this setup we trained on two (partially annotated) kidney volumes and used the trained network to segment the third volume. We report the result on all 3 possible combinations of training and test volumes. Table~\ref{tab:cv_results_naive} summarizes the IoU as in the previous section over all annotated 2D slices of the left out volume. In this experiment BN also improves the result, except for the third setting, where it was counterproductive. We think that the large differences in the data sets are responsible for this effect. The typical use case for the fully-automated segmentation will work on much larger sample sizes, where the same number of sparse labels could be easily distributed over much more data sets to obtain a more representative training data set.

\section{Conclusion}\label{references}

We have introduced an end-to-end learning method that semi-automatically and fully-automatically segments a 3D volume from a sparse annotation. It offers an accurate segmentation for the highly variable structures of the \species{Xenopus} kidney. We achieve an average IoU of 0.863 in 3-fold cross validation experiments for the semi-automated setup. In a fully-automated setup we demonstrate the performance gain of the 3D architecture to 
an equivalent 2D implementation. The network is trained from scratch, and it is not optimized in any way for this application. We expect that it will be applicable to many other biomedical volumetric segmentation tasks. Its implementation is provided as OpenSource.

\subsubsection*{Acknowledgments.} We thank the DFG (EXC 294 and CRC-1140 KIDGEM Project Z02 and B07) for supporting this work. Ahmed Abdulkadir acknowledges funding by the grant KF3223201LW3 of the ZIM (Zentrales Innovationsprogramm Mittelstand). Soeren S. Lienkamp acknowledges funding from DFG (Emmy Noether-Programm). We also thank Elitsa Goykovka for the useful annotations and Alena Sammarco for the excellent technical assistance in imaging.

\bibliographystyle{splncs03}
\bibliography{references}

\end{document}